\documentclass[sigconf]{acmart}
\AtBeginDocument{%
  }

\setcopyright{acmlicensed}
\copyrightyear{2018}
\acmYear{2018}
\acmDOI{XXXXXXX.XXXXXXX}
\acmConference[Conference acronym 'XX]{Make sure to enter the correct
  conference title from your rights confirmation email}{June 03--05,
  2018}{Woodstock, NY}
\acmISBN{978-1-4503-XXXX-X/2018/06}

\usepackage{graphicx} 		
\usepackage{subcaption}		

\usepackage{stfloats}
\usepackage{booktabs}   
\usepackage{multirow}

\usepackage{amssymb}  
\usepackage{array}       
\usepackage{makecell}    

\usepackage{colortbl}
\usepackage{marvosym}

\newcommand{\eg}{\emph{e.g.}}
\newcommand{\ie}{\emph{i.e.}}
\begin{document}


\title{
\raisebox{-0.25em}{\includegraphics[height=1.2em]{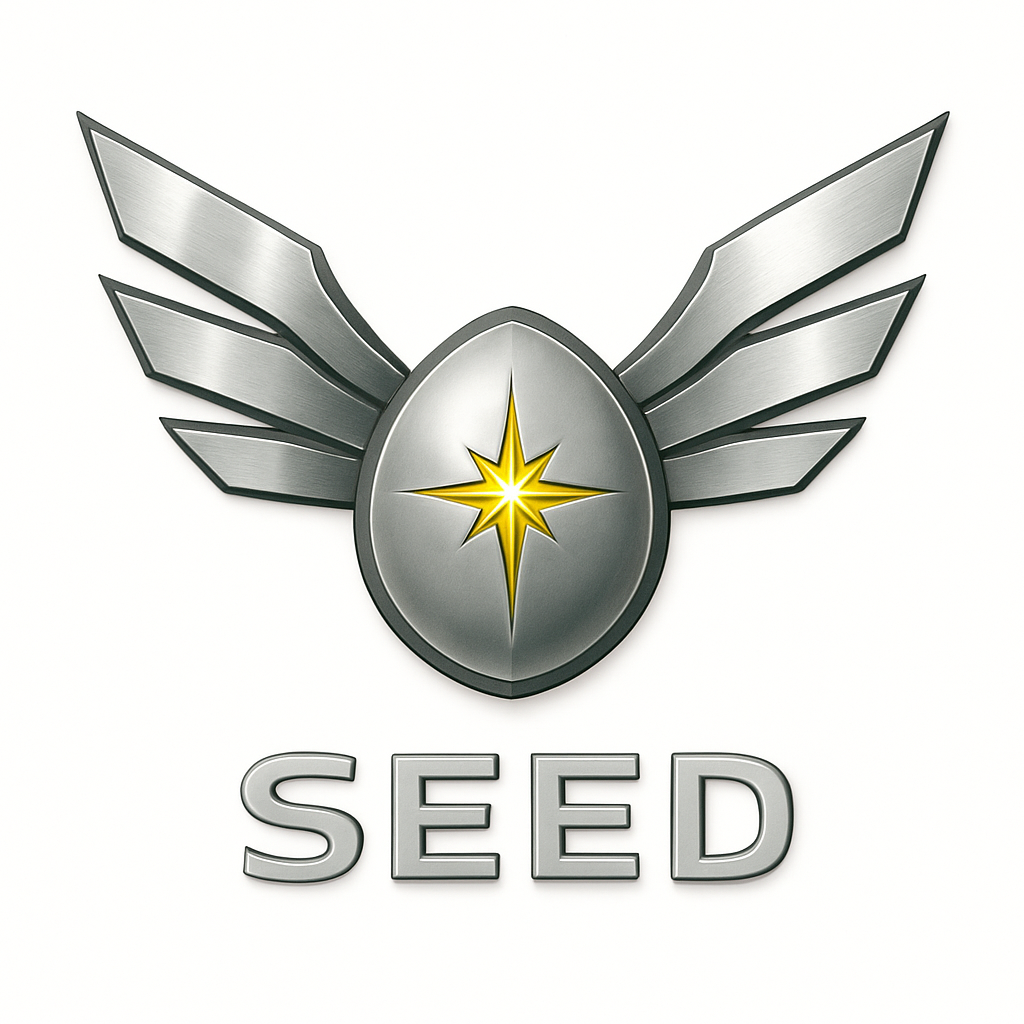}}SEED: A Benchmark Dataset for Sequential Facial Attribute Editing with Diffusion Models}

\author{Yule Zhu} 
\affiliation{
  \institution{Huazhong University of Science and Technology}
  \city{Wuhan}
  \country{China}
  }
\email{zyl_1037@hust.edu.cn}

\author{Ping Liu \Letter}\authornote{\Letter~denotes co-corresponding author.}
\affiliation{%
  \institution{University of Nevada Reno}
  \city{Reno, NV}
  \country{USA}}
\email{pino.pingliu@gmail.com}

\author{Zhedong Zheng} 
\affiliation{
  \institution{University of Macau}
  \city{Macau}
  \country{China}
  }
\email{zhedongzheng@um.edu.mo}

\author{Wei Liu~\Letter}
\affiliation{%
  \institution{Huazhong University of Science and Technology}
  \city{Wuhan}
  \country{China}
}
\email{liuwei@hust.edu.cn}






\renewcommand\thesubfigure{(\alph{subfigure})} 
\captionsetup[sub]{
	labelformat=simple
}
\captionsetup[subtable]{labelformat=simple, labelsep=space}
\renewcommand{\thesubtable}{(\alph{subtable})} 

\begin{abstract}
Diffusion models have recently enabled precise and photorealistic facial editing across a wide range of semantic attributes. 
Beyond single-step modifications, a growing class of applications now demands the ability to analyze and track sequences of progressive edits, such as stepwise changes to hair, makeup, or accessories. 
{However, sequential editing introduces significant challenges in edit attribution and detection robustness, further complicated by the lack of large-scale, finely annotated benchmarks tailored explicitly for this task.} 
We introduce \textbf{SEED}, a large-scale \underline{S}equentially \underline{E}dited fac\underline{E} \underline{D}ataset constructed via state-of-the-art diffusion models. 
SEED contains over 90,000 facial images with one to four sequential attribute modifications, generated using diverse diffusion-based editing pipelines (LEdits, SDXL, SD3). 
Each image is annotated with detailed edit sequences, attribute masks, and prompts, {facilitating research on sequential edit tracking, visual provenance analysis, and manipulation robustness assessment.} 
To benchmark this task, we propose \textbf{FAITH}, a frequency-aware transformer-based model that incorporates high-frequency cues to enhance sensitivity to subtle sequential changes. 
{Comprehensive experiments, including systematic comparisons of multiple frequency-domain methods, demonstrate the effectiveness of FAITH and the unique challenges posed by SEED.}
SEED offers a challenging and flexible resource for studying progressive diffusion-based edits at scale. 
Dataset and code will be publicly released at: \url{https://github.com/Zeus1037/SEED}.
\end{abstract}




\received{20 February 2007}
\received[revised]{12 March 2009}
\received[accepted]{5 June 2009}

\maketitle

\section{Introduction}
Recent advances in generative modeling {\cite{vae, gan, dm}} have significantly improved the realism and controllability of facial image editing.
Among these techniques, diffusion models (DMs) {\cite{ddim, ddpm, Rombach_2022_CVPR(SD)}} have emerged as a dominant paradigm due to their ability to produce photorealistic outputs with precise, localized, and semantically meaningful edits.
Compared to traditional generative approaches, diffusion models offer significantly improved visual realism and finer control over edits, enabling users to \textit{sequentially} modify facial attributes, such as adjusting hairstyle, then applying virtual makeup{~\cite{ textguidedeyeglasses, stablemakeup, huang2021real}}, and subsequently adding realistic eyeglasses ~\cite{unsupervisedeyeglasses}, with minimal visual artifacts and higher fidelity. 
Such sequential editing workflows are particularly useful in applications like interactive virtual try-on and personalized content creation.

However, the increasing sophistication of \textit{sequential facial attribute editing} presents substantial new challenges in visual analysis and authenticity verification. 
Sequential manipulations create subtle, cumulative changes across multiple attributes, complicating the accurate identification of which attributes have been altered, the sequence of edits, and the specific editing methodologies applied. 
Despite these pressing challenges, research into sequential editing detection remains significantly under-explored, {primarily due to the absence of sufficiently large-scale and fine-grained annotated benchmarks tailored for diffusion-generated content.}
The limited available dataset, SeqDeepFake~\cite{shao2022detecting(SeqFakeFormer)}, contains fewer than 50K GAN-generated images and provides {relatively} coarse-grained annotations, hindering its ability {for benchmarking contemporary diffusion-based sequential editing methods (Table~\ref{Sequential Deepfake Dataset}).}
Moreover, models trained solely on GAN-generated images exhibit severe performance degradation when applied to diffusion-generated edits~\cite{ricker2024towards_the_detection_of_DM_deepfakes,Tantaru_2024_WACV}. 
{These gaps highlight the critical need for a large-scale, finely annotated dataset and robust analytical methods tailored to the challenges of diffusion-generated sequential edits while maintaining broad applicability.}

\begin{figure*}[htbp]
  \centering
  \includegraphics[width=0.98\linewidth]{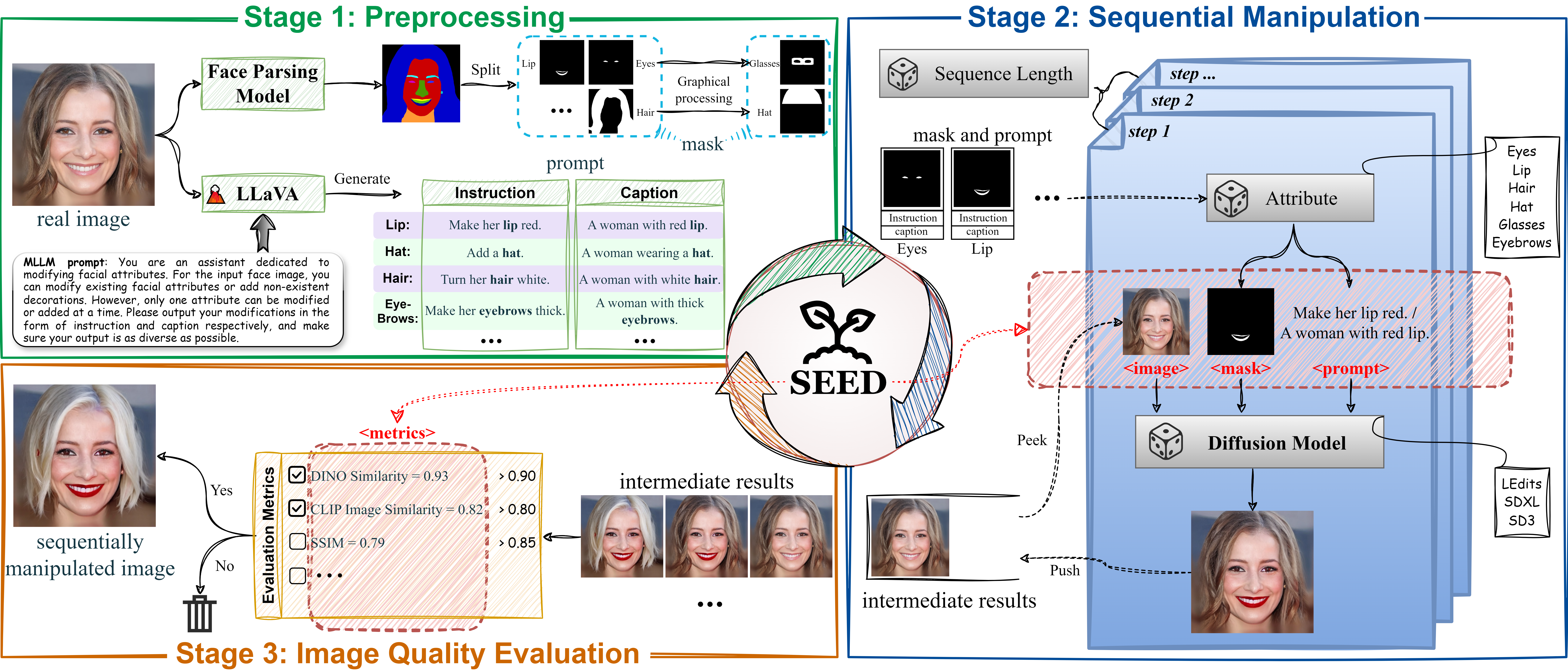}
\caption{Overview of the SEED Dataset Generation Pipeline. Facial images first undergo mask and prompt generation using Face Parsing and LLaVA. 
Next, editing sequences and diffusion models (LEdits, SDXL, UltraEdit) are randomly selected at each step for sequential editing. 
Finally, edited images are rigorously quality-assessed, and detailed annotations—including images, masks, prompts, and quality metrics—are systematically recorded, ensuring dataset realism and diversity.}
  \Description{Dataset generation pipeline.}
  \label{Dataset generation pipeline}
\end{figure*}

To bridge this gap, we introduce \textbf{SEED}, the first large-scale \underline{S}equentially \underline{E}dited fac\underline{E} \underline{D}ataset created entirely using state-of-the-art diffusion models. 
SEED comprises  $91,526$ facial images, sequentially edited across one to four attributes. 
Our data generation pipeline integrates three advanced diffusion-based frameworks, LEdits, SDXL, and SD3, with automatic facial parsing, targeted prompt generation, and localized mask-based editing.
{As illustrated in Figure~\ref{Dataset generation pipeline}, SEED provides extensive annotations, including exact editing sequences, diffusion models used, specific prompts, and precise masks, facilitating detailed downstream analyzes like editing sequence prediction and manipulation region localization.}

{Building upon SEED, we further propose \textbf{FAITH}, a Frequency-Aware Identification Transformer specifically designed for hierarchical editing sequence detection.}
Unlike traditional spatial-only methods~\cite{shao2022detecting(SeqFakeFormer)}, {FAITH integrates high-frequency domain information into the Transformer's cross-attention mechanism, providing explicit spatial guidance to better capture subtle editing artifacts.}
{Notably, FAITH represents the first attempt to explore frequency-domain cues in sequential facial editing detection.}
We systematically compare multiple frequency transformations—Discrete Cosine Transform (DCT)~\cite{dct}, Fast Fourier Transform (FFT)~\cite{dsp}, and Discrete Wavelet Transform (DWT)~\cite{dsp}—demonstrating that DWT achieves superior performance due to its multi-scale representation and effective capture of directional textures and edges introduced by diffusion-based editing. In summary, our contributions are threefold as follows:
\begin{itemize}
    \item We present the first large-scale diffusion-based sequential facial editing dataset, {SEED}, comprising $91,526$ images with fine-grained annotations, including precise editing sequences, textual prompts, and manipulation masks, effectively addressing critical limitations of existing datasets.

    \item We propose {\color{black}{FAITH}}, a novel frequency-aware sequential editing analysis model that integrates high-frequency domain features, improving sensitivity to subtle sequential edits.

    \item We perform extensive benchmark experiments on SEED, systematically comparing frequency-domain transformations (DCT, FFT, DWT), highlighting both the inherent challenges of the dataset and the efficacy of frequency-guided analysis.
\end{itemize}

\section{Related Work}
\noindent\textbf{Facial Manipulation Datasets.}
Single-step facial manipulations typically involve isolated edits, such as face swapping or attribute modifications. 
Driven by advances in generative modeling, extensive datasets based on GANs~\cite{10.1145/3422622(GAN)} and diffusion models~\cite{ddim, ddpm, Rombach_2022_CVPR(SD)} have emerged, including ForgeryNet~\cite{he2021forgerynet(forgerynet)}, OpenForensics~\cite{Le_2021_ICCV(openforensics)}, DF-Platter~\cite{Narayan_2023_CVPR(df-platter)}, and the recently introduced DiffusionDB~\cite{wang2022diffusiondb(DiffusionDB)}, containing 14 million Stable Diffusion-generated images. 
Other relevant datasets like DeepFakeFace~\cite{song2023robustness(DeepFakeFace)} and DiffusionFace~\cite{chen2024diffusionface(DiffusionFace)} similarly focus only on isolated facial edits.
However, real-world scenarios frequently involve multiple subtle and cumulative edits applied sequentially, presenting greater challenges for detection and analysis. 
{Despite the practical importance of sequential facial editing}, datasets explicitly addressing sequential facial editing remain sparse. 
SeqDeepFake~\cite{shao2022detecting(SeqFakeFormer)}, one of the few available sequential datasets, {consists mainly of GAN-generated images, restricting its ability to represent the complexity and realism of contemporary diffusion-generated edits.}
{To bridge this gap, our SEED dataset introduces large-scale sequential edits generated by state-of-the-art diffusion models, providing fine-grained annotations to significantly advance research in sequential facial editing detection and analysis.}

\noindent\textbf{Facial Manipulation Detection.}
Detection methods for facial manipulations generally target visual inconsistencies \cite{tan2024data-independent_operator,Nguyen_2024_CVPR,Tan_2024_CVPR}, frequency-domain irregularities \cite{Corvi_2023_CVPR,chen2024singlesimplepatchneed,freqnet}, and temporal artifacts \cite{Liu_2023_WACV,Choi_2024_CVPR,zhang2025learning}. 
However, images generated by diffusion models present new challenges due to their increased realism \cite{ricker2024towards_the_detection_of_DM_deepfakes,Tantaru_2024_WACV, beyondfaceswapping}. Recent studies by Wang \textit{et al.}~~\cite{Wang_2023_ICCV(DIRE)} and Liu \textit{et al.}~~\cite{liu2024BMSB} utilize reconstruction errors and spatial-frequency analyses to improve detection of diffusion-generated images. For a comprehensive overview, readers may refer to recent surveys \cite{liu2024robust,heidari2024deepfake,liu2024evolvingsinglemodalmultimodalfacial}.
{Sequential editing detection poses additional challenges as it requires precise identification of specific editing sequences and manipulated attributes.}
Existing sequential detection methods \cite{shao2022detecting(SeqFakeFormer),shao2023robust(SeqFakeFormer++),xia2024mmnet(MMNet),hong2024contrastive,Li_2025_WACV} mostly target GAN-generated images, thus exhibiting limited generalization to diffusion-generated sequences. 
To mitigate this critical limitation, we propose FAITH, a novel frequency-aware transformer framework explicitly designed for diffusion-generated sequential facial editing scenarios, significantly enhancing sensitivity and detection accuracy through integration of high-frequency domain features.

\begin{table}[t!]
\footnotesize\caption{Comparison of Sequential Facial Editing Datasets}
\label{Sequential Deepfake Dataset}
\begin{tabular}{cccc}
\hline
\rowcolor{gray!20} 
\cellcolor{gray!20}                                      & \multicolumn{2}{c}{\cellcolor{gray!20} SeqDeepFake}                                                     & \cellcolor{gray!20}                                                                                                             \\ \cline{2-3}
\rowcolor{gray!20} 
\multirow{-2}{*}{\cellcolor{gray!20} \textbf{Dataset}}                & components                        & attributes                                                              & \multirow{-2}{*}{\cellcolor{gray!20} \textbf{SEED(Ours)}}                                                                        \\ \hline
 \textbf{Size}                                                     & 35,166       & 49,920                   & 91,526                   \\
\rowcolor[HTML]{ececec} 
\textbf{\#Attr.}                                                  & 5 &  5                       & \textbf{6}                             \\
\textbf{\#Seq.}                                                   & 28          & 26                      & 200                    \\
\rowcolor[HTML]{ececec} 
\textbf{\begin{tabular}[c]{@{}c@{}}Editing\\ Methods\end{tabular}} & StyleMapGAN~\cite{Kim_2021_CVPR(stylemapgan)} & \begin{tabular}[c]{@{}c@{}}StyleGAN~\cite{FFHQ} +\\  Talk-to-edit~\cite{talktoedit}\end{tabular} & \begin{tabular}[c]{@{}c@{}}LEdits~\cite{tsaban2023ledits(LEdits)}\\ SDXL~\cite{podell2023sdxl(SDXL)}\\ SD3~\cite{zhao2024ultraeditinstructionbasedfinegrainedimage}\end{tabular} \\
\textbf{\begin{tabular}[c]{@{}c@{}}Supporting\\ Tasks\end{tabular}} & \multicolumn{2}{c}{\begin{tabular}[c]{@{}c@{}} forgery detection\\ sequence prediction\end{tabular}} & \begin{tabular}[c]{@{}c@{}} forgery detection\\ sequence prediction\\ spatial localization\end{tabular} \\ \hline
\end{tabular}
\end{table}

\section{SEED: Sequential Facial Attribute Editing Benchmark}
In this work, we introduce \textbf{SEED}, the first large-scale sequential facial attribute editing dataset generated by diffusion models, comprising 91,526 edited images derived from {FFHQ~\cite{FFHQ} and CelebAMask-HQ~\cite{CelebAMask-HQ}}.
Each image undergoes 1 to 4 sequential editing steps, involving basic attributes (\eg, eyes, lips, hair) or accessories (\eg, glasses, hats). 
Random combinations of attributes ensure that SEED maintains exceptional diversity and generalizability. 
Moreover, our carefully designed data generation pipeline preserves realistic visual effects throughout sequential manipulations.

\subsection{Diffusion-based Sequential Editing Pipeline}
As illustrated in Figure~\ref{Dataset generation pipeline}, our pipeline includes three stages: preprocessing, sequential manipulation, and image quality evaluation. 
The structured design facilitates efficient and scalable generation of diverse sequential edits from real facial images. 
Furthermore, the pipeline allows easy integration of additional editing methods, supporting the creation of diverse and complex scenarios.

\noindent\textbf{Preprocessing Stage.}
This stage generates masks identifying edited regions and textual prompts guiding diffusion-based editing, crucial for automatic and disentangled manipulations. 
Face masks are generated using the CelebAMask-HQ face parsing model~\cite{CelebAMask-HQ}, which precisely segments facial components such as eyes and hair. 
Accurate segmentation ensures that each editing step modifies only the intended facial area, effectively minimizing interference between consecutive editing operations.
To synthesize textual prompts, we employ LLaVA~\cite{liu2023llava} to produce multiple attribute-specific prompts, randomly selecting one for each editing step. 
{
Prompts follow two primary template formats, explicitly designed based on editing methods: instruction-based prompts (\eg, "Make his/her hair curly") and description-based prompts (\eg, "A man/woman wearing a black hat").
This diverse and realistic prompt generation strategy helps replicate practical user interactions during sequential facial attribute editing.
}
For additive edits such as "Add glasses" or "Add a hat," we apply simple yet effective graphical techniques, such as dilating the eye mask to approximate a glasses mask or selecting the upper hair region as a proxy for hat placement.
Although these masks are approximate, they have proven sufficiently effective for realistic data generation.

\noindent\textbf{Sequential Manipulation Stage.} 
In this stage, the pipeline sequentially edits facial images according to randomly generated attribute sequences. 
Specifically, we first determine the length of the editing sequence by randomly selecting a number between 1 and 4. 
For each editing step, an attribute is randomly chosen from an expanded candidate set, including fundamental facial attributes (eyes, lips, hair, eyebrows) as well as additive decorative edits (\eg, glasses or hats). 
The randomness in attribute selection significantly enhances the diversity of manipulation sequences and reduces potential biases associated with any specific attribute combination, thus making SEED more representative of diverse real-world editing scenarios. 
The pipeline then retrieves the corresponding mask and textual prompt previously generated in the preprocessing stage for each selected attribute.
Subsequently, the input image (initially the real face, subsequently each edited result), along with its respective mask and prompt, is processed using one of three diffusion-based editing methods, \ie, LEdits~\cite{tsaban2023ledits(LEdits)}, SDXL~\cite{podell2023sdxl(SDXL)}, and Stable Diffusion 3 fine-tuned with UltraEdit~\cite{zhao2024ultraeditinstructionbasedfinegrainedimage}. 
The specific editing method for each step is also randomly selected to further enhance dataset diversity and minimize biases associated with a single editing model. 
These chosen methods excel at producing disentangled attribute modifications and precisely following textual instructions. The output from each editing step then serves as input for subsequent steps, ensuring coherent and realistic sequential edits. 
Moreover, our pipeline is designed with modularity, allowing easy integration of additional advanced image editing methods, thereby providing excellent scalability.
Finally, all sequentially manipulated images are passed on to the next stage for image quality evaluation.

\noindent\textbf{Image Quality Evaluation Stage.}
Edited images undergo comprehensive quality assessments using multiple complementary metrics, including DINO similarity~\cite{zhang2022dino}, CLIP image similarity~\cite{clip}, and Structural Similarity Index (SSIM)~\cite{SSIM}. 
DINO and CLIP evaluate semantic coherence and visual consistency, while SSIM measures structural integrity at pixel-level. 
Only images exceeding predefined quality thresholds—ensuring visual fidelity—are retained. 
Quality thresholds are adjustable, allowing flexible control based on specific research needs or practical requirements. 
Through this rigorous and comprehensive evaluation strategy, we guarantee that SEED provides high-quality and visually realistic data samples, as exemplified in Supplementary.

\begin{figure}[t]
    \centering
    \includegraphics[width=\linewidth]{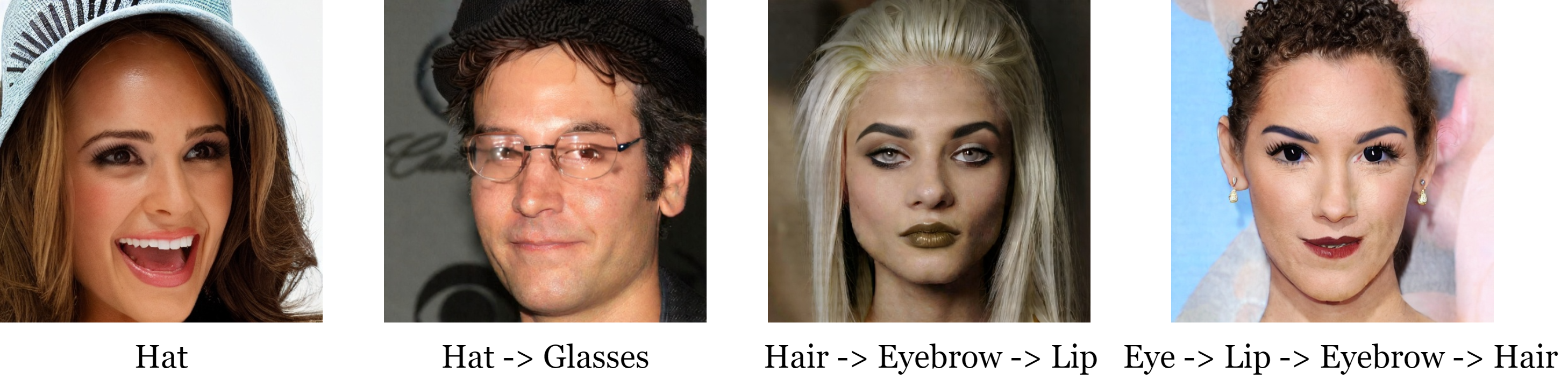}
    \caption{Samples from SEED demonstrating sequential facial attribute edits. Modified attributes and editing order are annotated below each image.}
    \Description{Samples of SEED.}
    \label{Samples of SEED}
\end{figure}
\subsection{Dataset Statistics}
{SEED consists of 91,526 facial images sourced from FFHQ~\cite{FFHQ} and CelebAMask-HQ~\cite{CelebAMask-HQ}, ensuring diverse representation across ages, ethnicities, and poses.} 
{Each image (512$\times$512 resolution) undergoes sequential editing across one to four randomly chosen attributes, with sequence lengths distributed relatively evenly (29.91\%, 26.21\%, 21.88\%, and 22.00\%).}
{Attribute selection and diffusion model choices reflect common real-world editing preferences.}
{Comprehensive annotations—including masks, textual prompts, diffusion models, and quality metrics—are provided, supporting reproducibility and enabling various downstream tasks such as manipulation detection and sequence prediction, as exemplified in Supplementary.}



\noindent\textbf{Distinctive Features of SEED.} Compared to GAN-based datasets like SeqDeepFake~\cite{shao2022detecting(SeqFakeFormer)}, SEED employs advanced diffusion models (LEdits, SDXL, UltraEdit) for more realistic edits, diverse attributes, and finer annotations. 
{Additionally}, SEED notably expands attributes beyond basic facial features to include decorative elements (\eg, glasses, hats), closely aligning with practical scenarios such as virtual try-on. 
{Moreover}, editing sequences and diffusion methods are randomized at each step, further enhancing dataset diversity and reducing single-model biases. 
{Overall}, SEED provides a rigorous benchmark specifically tailored for sequential editing analysis using state-of-the-art diffusion-based approaches.

\begin{figure*}[htbp]
  \centering
  \includegraphics[width=0.90\linewidth]{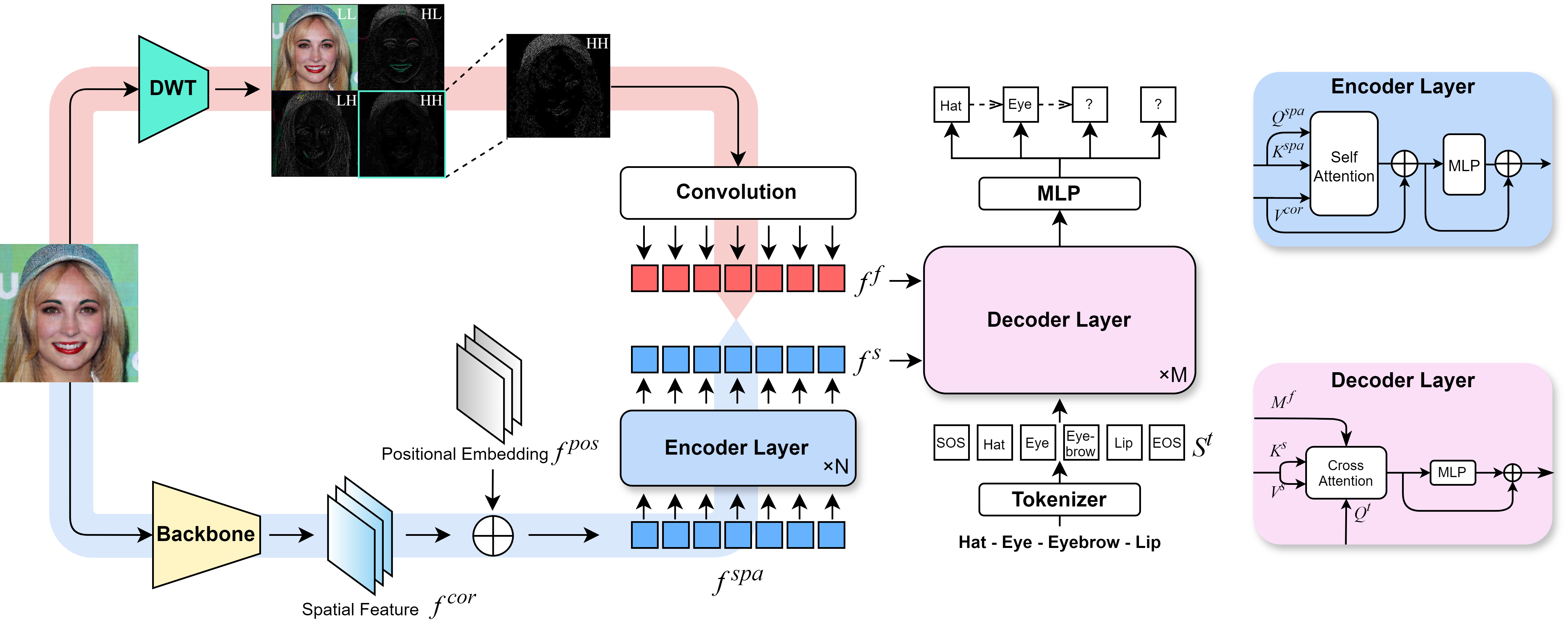}
  \caption{Overview of the proposed FAITH architecture. The model explicitly integrates high-frequency domain information extracted by Discrete Wavelet Transform (DWT) and spatial-domain features via a Transformer-based encoder-decoder structure, significantly enhancing sequential facial editing detection accuracy.}
  \Description{Sequential deepfake detection model.}
  \label{Sequential deepfake detection model}
\end{figure*}

\section{FAITH: Frequency-Aware Identification Transformer for Sequential Editing Detection}
{Existing sequential facial editing detection models, such as SeqFakeFormer~\cite{shao2022detecting(SeqFakeFormer)}, primarily rely on spatial-domain features, resulting in limited effectiveness when handling subtle and realistic edits produced by diffusion models~\cite{ricker2024towards_the_detection_of_DM_deepfakes}.}  
{Moreover, their spatial attention maps, derived solely from label embeddings, fail to accurately highlight manipulated regions.}  
{To address these challenges, we propose a novel frequency-aware transformer model, {FAITH}, specifically designed as a robust baseline for sequential editing detection on the SEED dataset.}
{As illustrated in Figure~\ref{Sequential deepfake detection model}, coarse spatial features ($f^{\text{cor}}$) are extracted by the CNN backbone, refined into spatial sequential features ($f^\text{{spa}}$), and further processed by a Transformer encoder to produce refined spatial features ($f^\text{{s}}$). 
Frequency-domain features ($f^\text{{f}}$), specifically the HH (high-high) sub-band from DWT, capturing high-frequency details in both horizontal and vertical directions, serve as auxiliary spatial guidance ($M^\text{{f}}$) during decoding to enhance detection sensitivity.}

\noindent\textbf{Spatial Feature Extraction.}
{To effectively capture subtle editing traces, FAITH explicitly extracts both spatial and frequency-domain features from the input image.}  
{Initially, a CNN backbone extracts coarse-grained spatial features $f^{\text{cor}}$ from the input face image $\mathbf{x}$.}  
Then, positional embeddings are added, and these features are flattened into a sequential form $f^{\text{spa}}$:
\begin{equation}
    f^{\text{spa}} = \text{Flatten}(\text{CNN}(\mathbf{x}) + f^{\text{pos}}).
\end{equation}

{These sequential spatial features are subsequently refined through an N-layer Transformer encoder, leveraging a self-attention mechanism where $f^{\text{spa}}$ serves as queries and keys, and $f^{\text{cor}}$ as values:}
\begin{gather}
    f^{\text{mid}} = \text{SoftMax}(QK^{T}/\sqrt{d})V + V, \\
    f^{\text{s}} = f^{\text{mid}} + \text{MLP}(f^{\text{mid}}),
\end{gather}
where $d$ denotes the dimension of queries and keys.

\noindent\textbf{Frequency-domain Feature Extraction.}
{Since high-frequency image components effectively reveal subtle manipulation traces, FAITH explicitly leverages discrete wavelet transform (DWT) to enhance detection sensitivity.}  
{Specifically, we extract the high-frequency HH subband via DWT and employ convolutional layers to encode spatial inconsistencies arising from editing operations:}
\begin{equation}
    f^{\text{f}} = \text{Conv}(\text{HH}(\text{DWT}(\mathbf{x}))),
\end{equation}
{where HH($\cdot$) represents the extraction of the HH component after wavelet transformation. Integrating these frequency features alongside spatial representations significantly enhances FAITH’s robustness against subtle sequential edits.}

\noindent\textbf{Cross-attention based Sequential Edit Prediction}
{To precisely model relationships between extracted image features and sequential editing labels, FAITH employs a decoder with cross-attention mechanisms guided by high-frequency features.}  
Specifically, we tokenize each edited attribute (\eg, Hat, Eye, Eyebrow, Lip), and insert special tokens (SOS/EOS) to indicate editing sequence boundaries, forming a token sequence $S^{\text{t}}$.
{In the Transformer decoder, we calculate cross-attention between the tokenized manipulation sequences (using $S^{\text{t}}$ as queries) and refined spatial features ($f^{\text{s}}$ as keys and values).}  
{Critically, the frequency-domain features $f^{\text{f}}$ serve as auxiliary spatial guidance, enabling the model to concentrate on regions likely modified by editing:}
\begin{gather}
    f^{\text{mid}'}_{i} = \text{SoftMax}(QK^{T}/\sqrt{d} + M^{\text{f}})V, \\
    f^{\text{dec}}_i = f^{\text{mid}'}_{i} + \text{MLP}(f^{\text{mid}'}_{i}), \\
    Attr_i = \text{MLP}(f^{\text{dec}}_i),
\end{gather}
{where $M^{\text{f}}$ is derived from high-frequency features $f^{\text{f}}$, and queries $Q$ are generated from sequence tokens $S^{\text{t}}$ after self-attention processing.}
{FAITH employs an auto-regressive decoding strategy guided by EOS tokens to sequentially predict editing attributes until termination.}  
The training procedure utilizes standard Cross-Entropy Loss, supervised by ground-truth attribute sequences.

\section{Experiment}

\subsection{Implementation Details}

\textbf{Balanced Partition of SEED.} Since detection difficulty increases with longer editing sequences, uneven distributions could bias evaluation. 
To address this issue, we construct a carefully designed, sequence-length-balanced partition of SEED. 
Specifically, we randomly select 20,000 samples for each editing sequence length from 1 to 4 edits and additionally include 20,000 unedited real images with sequence length of 0 from FFHQ and CelebAMask-HQ. 
This results in a dataset of 100,000 images equally distributed across sequence lengths, each accounting for 20\% of the total. 
The partition is further split into training, validation, and test sets with a ratio of 8:1:1, ensuring a fair evaluation.

\noindent\textbf{Evaluation Metrics.} 
{We comprehensively evaluate sequential editing analysis models using three complementary metrics.}
{{Fixed-Acc}} explicitly compares each predicted attribute against its ground-truth label at fixed positions (maximum length of 4), marking positions without edits as "no manipulation".
{{Adaptive-Acc}} measures accuracy adaptively by comparing predictions and labels only up to the length of the shorter sequence between ground truth and prediction, avoiding bias from unedited positions.
{Finally, the rigorous sequence-level metric, {Full-Acc}, considers predictions correct only if the predicted attribute sequence exactly matches the ground-truth, directly assessing the model’s capability to precisely recover the editing sequence.}

\noindent\textbf{Training Details.}
{We first detail the implementation settings for training FAITH.}  
We adopt ResNet-50~\cite{resnet50}, pre-trained on ImageNet~\cite{imagenet}, as our spatial feature backbone.  
{The Transformer contains 2 encoder and 2 decoder layers, each with 4 attention heads.}  
We employ learning rate warm-up for 20 epochs followed by step-wise decay every 50 epochs, totaling 170 epochs, starting at $1\times10^{-3}$ for Transformer and $1\times10^{-4}$ for ResNet.  
{Batch size is 40, balancing varying-length sequences per batch.}  
Experiments use PyTorch and three NVIDIA 3090 GPUs, with SAM optimizer~\cite{samoptimizer} for improved generalization.
{For comparison, we adapt state-of-the-art facial manipulation detectors~\cite{freqnet,locate_and_verify,exposing_the_deception} from binary to multi-class (200 unique sequences) to assess their sequential detection capabilities.}  
{We also include SeqFakeFormer~\cite{shao2022detecting(SeqFakeFormer)}, a representative method relying solely on spatial features, clearly demonstrating FAITH's advantage in incorporating frequency-domain information.}

\begin{table*}[htbp]
\centering
\footnotesize
\caption{
Performance comparison of manipulation detection methods on SEED, including traditional methods~\cite{locate_and_verify,freqnet,exposing_the_deception}, baseline SeqFakeFormer~\cite{shao2022detecting(SeqFakeFormer)}, and FAITH enhanced by frequency-domain features (DCT, FFT, DWT). Best average results (in bold) show frequency-domain features, particularly DWT HH, significantly improve detection of longer editing sequences.
}
\label{methods comparison}
\begin{tabular}{cc|ccc|cccc}
\toprule 
\multicolumn{2}{c|}{\textbf{Methods}} & {Locate and verify~\cite{locate_and_verify}} & {FreqNet~\cite{freqnet}} & {Exposing the deception~\cite{exposing_the_deception}} & {SeqFakeFormer~\cite{shao2022detecting(SeqFakeFormer)}} & \textbf{FAITH(DCT)} & \textbf{FAITH(FFT)} & \textbf{FAITH(DWT)} \\
\bottomrule
  & \textit{Fixed}   & 99.95    & 100  & 99.99  & 100    & 100   & 100   & 100   \\
0  & \textit{Adaptive}     & 99.80    & 100  & 99.95  & 100    & 100   & 100   & 100   \\
   & \textit{Full}    & 99.80    & 100  & 99.95  & 100    & 100   & 100   & 100   \\
\midrule
  & \textit{Fixed} & 97.32 & 93.77 & 93.25  & 98.00  & 98.10 & 98.29 & 98.24 \\
1   & \textit{Adaptive} & 89.50 & 79.01 & 86.61  & 92.11  & 92.48 & 93.21 & 93.01 \\
& \textit{Full} & 90.15 & 80.37 & 87.98 & 92.80  & 93.10 & 93.60 & 93.40 \\
\midrule
  & \textit{Fixed} & 78.12 & 74.01 & 68.01  & 89.50  & 89.76 & 89.70 & 89.69 \\
2 & \textit{Adaptive}  & 56.88 & 49.82 & 48.62  & 79.44  & 79.82 & 79.62 & 79.62 \\
   & \textit{Full} & 52.25 & 46.09 & 52.31  & 76.75  & 77.10 & 77.00 & 77.10 \\
\midrule
  & \textit{Fixed}  & 53.66 & 51.34   & 52.60  & 70.44  & 71.12 & 71.61 & 72.50 \\
3   & \textit{Adaptive}   & 35.53  & 32.54    & 40.62  & 61.56  & 62.21 & 62.67 & 63.84 \\
   & \textit{Full} & 22.45   & 25.57    & 26.58  & 48.55  & 49.65 & 49.25 & 50.35 \\
\midrule
  & \textit{Fixed}  & 28.45 & 26.93    & 34.20  & 50.14  & 49.51 & 49.14 & 48.93 \\
4   & \textit{Adaptive}  & 28.45   & 26.93    & 34.20  & 50.14  & 49.51 & 49.14 & 48.93 \\
   & \textit{Full}  & 5.70  & 10.92     & 7.20   & 24.55  & 22.95 & 23.05 & 23.35 \\
\midrule
 & \textit{Fixed} & 71.50 & 70.08    & 68.78  & 81.62  & 81.70 & 81.75 & \textbf{81.87} \\
 Avg.  & \textit{Adaptive}  & 48.72  & 48.27    & 50.80  & 66.97  & 67.02 & 67.03 & \textbf{67.26} \\
   & \textit{Full} & 54.07  & 52.59    & 54.80  & 68.53  & 68.56 & 68.58 & \textbf{68.84} \\
\bottomrule
\end{tabular}
\end{table*}

\subsection{Results and Analysis}
{To thoroughly verify the effectiveness of frequency-domain features in sequential facial editing detection, we comprehensively compare three frequency-domain transformations: Discrete Cosine Transform (DCT), Fast Fourier Transform (FFT), and Discrete Wavelet Transform (DWT).}
Detailed comparison results of the baseline and FAITH integrated with frequency-domain methods are presented in Table~\ref{methods comparison}. 
{We observe that traditional facial manipulation detection models~\cite{freqnet,locate_and_verify,exposing_the_deception} perform poorly when adapted to sequential editing scenarios, achieving an average Full-Acc just slightly above 50\%, notably lower than methods explicitly designed for sequential tasks.}
{Additionally, the detection accuracy consistently declines as the editing sequence length increases, indicating greater detection difficulty with longer editing sequences.}
By integrating frequency-domain features, FAITH significantly enhances performance across all three metrics.
{We particularly emphasize the improvement in Full-Acc, as accurately predicting entire editing sequences is more crucial than individual attributes.}
While DCT and FFT yield similar accuracy improvements, DWT demonstrates the highest performance, achieving a {0.30\%} improvement over the baseline.
{This superiority stems from the HH sub-band of DWT, capturing high-frequency components in both horizontal and vertical directions, thus retaining richer directional textures and edges compared to the non-directional detail captured by DCT and FFT.}

\begin{table*}[!ht]
\centering
\footnotesize
\caption{Robustness comparisons on the SEED test set under two scenarios: (a) JPEG compression (25\%, 50\%, 75\%) and (b) Gaussian noise (10\%, 15\%, 20\%). 
We evaluate non-sequential methods, the sequential approach (SeqFakeFormer), and our FAITH (DCT, FFT, DWT).
Best results (bold) show that DWT consistently achieves superior robustness in both cases.}
\label{robustness}
\begin{subtable}[t]{\textwidth}
\centering
\caption{}
\label{JPEG Compress}
\begin{tabular}{l *{3}{ccc}}
\toprule
\multirow{2}{*}{\textbf{Methods}} & 
\multicolumn{3}{c}{Compress -- 25\%} & 
\multicolumn{3}{c}{Compress -- 50\%} & 
\multicolumn{3}{c}{Compress -- 75\%} \\
\cmidrule(lr){2-4} \cmidrule(lr){5-7} \cmidrule(lr){8-10}
& \textit{Fixed} & \textit{Adaptive} & \textit{Full} & 
\textit{Fixed} & \textit{Adaptive} & \textit{Full} & 
\textit{Fixed} & \textit{Adaptive} & \textit{Full} \\
\midrule
Locate and verify~\cite{locate_and_verify} & 63.98 & 47.81 & 50.21 & 61.06 & 46.29 & 48.68  & 56.87 & 32.69 & 36.22 \\
FreqNet~\cite{freqnet} & 63.20 & 47.09 & 50.03 & 60.73 & 46.04 & 48.70 & 56.41 & 32.44 & 36.43   \\
Exposing the deception~\cite{exposing_the_deception} & 63.15 & 47.47 & 50.36 & 59.93 & 45.90 & 48.98 & 56.76 & 33.01 & 36.41  \\
SeqFakeFormer~\cite{shao2022detecting(SeqFakeFormer)} & 78.68 & 61.60 & 63.67 & 44.41 & 75.83 & 59.24 & 62.01 & 66.63 & 39.33   \\
\textbf{FAITH(DCT)} & 78.68 & 61.41 & 63.93  & 76.89 & 59.78 & 62.09 & 66.80 & 39.18 & 44.36  \\
\textbf{FAITH(FFT)} & 78.83 & 61.43 & 63.98 & 76.64 & 59.37 & 62.10 & 66.93 & 39.60 & 44.69   \\
\textbf{FAITH(DWT)} & \textbf{78.92} & \textbf{61.91} & \textbf{64.13} & \textbf{76.92} & \textbf{59.86} & \textbf{62.75} & \textbf{67.19} & \textbf{39.71} & \textbf{44.79}  \\
\bottomrule
\end{tabular}
\end{subtable}

\begin{subtable}[t]{\textwidth}
\centering
\footnotesize
\caption{}
\label{add noise}
\begin{tabular}{l *{3}{ccc}}
\toprule
\multirow{2}{*}{\textbf{Methods}} & 
\multicolumn{3}{c}{Noise intensity -- 10\%} & 
\multicolumn{3}{c}{Noise intensity -- 15\%} & 
\multicolumn{3}{c}{Noise intensity -- 20\%} \\
\cmidrule(lr){2-4} \cmidrule(lr){5-7} \cmidrule(lr){8-10}
& \textit{Fixed} & \textit{Adaptive} & \textit{Full} & 
\textit{Fixed} & \textit{Adaptive} & \textit{Full} & 
\textit{Fixed} & \textit{Adaptive} & \textit{Full} \\
\midrule
Locate and verify~\cite{locate_and_verify} & 49.23 & 25.66 & 29.46 & 43.09 & 15.05 & 18.97 & 40.72 & 8.12 & 15.81 \\
FreqNet~\cite{freqnet} & 48.88 & 25.38 & 29.03 & 42.84 & 15.78 & 18.96 & 40.36 & 8.26 & 15.82 \\
Exposing the deception~\cite{exposing_the_deception} & 49.32 & 25.76 & 30.03 & 42.98 & 15.80 & 18.99 & 40.53 & 8.49 & 16.22 \\
SeqFakeFormer~\cite{shao2022detecting(SeqFakeFormer)}          & 62.39 & 31.62 & 34.59 & 55.61 & 19.28 & 23.40 & 51.21 & 11.31 & 20.27 \\
\textbf{FAITH(DCT)}    & 62.68 & \textbf{32.16} & \textbf{35.52} & 55.71 & 19.48 & 23.59 & 51.32 & 11.49 & 20.26 \\
\textbf{FAITH(FFT)}    & 62.91 & 31.76 & 34.37 & 55.28 & 18.95 & 23.06 & 51.51 & 11.83 & 20.39 \\
\textbf{FAITH(DWT)}    & \textbf{63.07} & 31.88 & 34.69 & \textbf{55.84} & \textbf{19.60} & \textbf{24.38} & \textbf{51.70} & \textbf{12.19} & \textbf{20.83} \\
\bottomrule
\end{tabular}
\end{subtable}
\end{table*}

\subsection{Robustness Study}
{Considering real-world image transmission scenarios involving compression or noise, we further evaluate model robustness against these post-processing conditions.}
We first perform JPEG compression at different ratios (25\%, 50\%, and 75\%) on the test set; results are summarized in Table~\ref{JPEG Compress}.
{As expected, accuracy significantly decreases as compression intensifies, with a dramatic drop at 75\% compression.}
However, our frequency-aware method consistently outperforms the baseline across all compression levels, highlighting the advantage of integrating high-frequency domain information.
{Additionally}, we add Gaussian noise at varying intensities (10\%, 15\%, and 20\%) to evaluate noise robustness; results are shown in Table~\ref{add noise}.
{Gaussian noise negatively impacts both spatial and frequency-domain information, reducing detection accuracy across all methods.}
Notably, the baseline relying solely on spatial-domain features exhibits significant performance degradation, whereas our FAITH model, leveraging the DWT-based HH frequency component, demonstrates comparatively stronger robustness, although performance under severe noise remains challenging.


\section{Conclusion}
This paper introduces SEED, the first large-scale sequential facial editing dataset containing over $90,000$ images generated via advanced diffusion models, along with detailed annotations and editing metadata. 
We propose a novel frequency-aware transformer (FAITH) leveraging high-frequency features, significantly enhancing sequential editing detection, as validated by extensive experiments. 
Future work includes integrating additional advanced editing frameworks~\cite{dalva2024fluxspace,yin2025instructattribute} to further improve SEED’s realism, diversity, and benchmark applicability.
\bibliographystyle{ACM-Reference-Format}
\bibliography{sample-base}

\appendix









\end{document}